\documentclass[
    reprint,
    aps,
    prx,
    superscriptaddress,
]{revtex4-2}
\usepackage{graphicx}
\PassOptionsToPackage{dvipsnames}{xcolor}
\usepackage{amsmath,amssymb}
\usepackage{microtype}
\usepackage[utf8]{inputenc}
\usepackage[T1]{fontenc}
\usepackage{newtxtext}
\usepackage[smallerops]{newtxmath}
\usepackage{nicefrac}
\usepackage{hyperref}
\hypersetup{
    colorlinks=true,
    linkcolor=NavyBlue,
    citecolor=NavyBlue,
    urlcolor=NavyBlue
}

\begin{document}

\title{Stochastic Resetting Accelerates Reinforcement Learning Beyond Random Search}

\author{Jello Zhou}
\affiliation{Biophysics Program, Stanford University, Stanford, California 94301, USA}

\author{David J. Schwab}
\thanks{VN and DJS contributed equally to this work.}
\affiliation{Initiative for the Theoretical Sciences and Princeton-CUNY Center for the Physics of Biological Function, The Graduate Center, CUNY, New York, NY 10016, USA}

\author{Vudtiwat Ngampruetikorn}
\thanks{VN and DJS contributed equally to this work.}
\affiliation{School of Physics, University of Sydney, Sydney, NSW 2006, Australia}
\affiliation{National Institute for Theory and Mathematics in Biology, Northwestern University and The University of Chicago, Chicago, IL 60611, USA}


\begin{abstract}
Stochastic resetting---intermittently returning a process to a fixed reference state---has emerged as an effective mechanism for optimizing first-passage properties.
Existing theory largely treats processes that search but do not learn: the searcher follows fixed dynamics, accumulating no knowledge between resets. 
Here we ask how stochastic resetting interacts with reinforcement learning, where the underlying dynamics adapt through experience.
In tabular grid environments, we find that resetting can accelerate learning even when it does not reduce the search time of a diffusive agent.
Our results reveal a distinct additional mechanism through which resetting speeds the propagation of reward information. 
We show that deterministic, sharp resetting accelerates learning more than the stochastic protocol but over a narrower range of reset rates.
In a continuous-state task with neural-network-based value approximation, we demonstrate that resetting speeds up learning when exploration is hard and rewards are sparse.
We argue further that, in the tabular tasks, resetting accelerates learning without altering the solution the agent ultimately reaches, unlike other techniques such as temporal discounting, which biases the optimal behavior.
Our results establish stochastic resetting as a simple, tunable mechanism for accelerating learning by shaping how experience accumulates, extending a canonical phenomenon of statistical mechanics to adaptive systems.
\end{abstract}

\maketitle

Restarting a process from a reference state is among the simplest forms of control. Foraging animals return to a home base between searches~\cite{orions79,lecheval24}, restart strategies accelerate combinatorial optimization~\cite{luby93,gomes98,montanari02} and kinetic proofreading exploits repeated reversion to an unbound state to achieve molecular specificity beyond equilibrium bounds~\cite{hopfield74,ninio75}. Despite such ubiquity, when and why restarts improve the performance of systems that learn from experience remains poorly understood.

Recent work has centered around developing an exactly solvable framework for analyzing such restarts~\cite{evans11,evans20,gupta22}. For a diffusive searcher on an unbounded domain, the first-passage time (FPT) to a target has infinite mean; resetting to a fixed position makes this mean finite with a minimum at a nonzero reset rate. The observation has grown into a broad theoretical program, including universality results~\cite{reuveni16, pal17} and extensions well beyond the original diffusive setting~\cite{kusmierz14,bressloff20,gupta20}.

Classical stochastic resetting theory has been developed primarily for static underlying processes---that is, the searcher does not adapt based on previous trajectories. Recent work has moved beyond that setting, including learning-induced localization in random search and adaptive frameworks for state- and time-dependent reset protocols~\cite{falcon17,keidar25}. Learning agents raise a complementary question: how does an external reset protocol change training when the agent keeps updating its behavior across experience?

Reinforcement learning (RL) offers a natural setting to study this interplay~\cite{sutton98}. An RL agent improves its policy by interacting with an environment and updating its value function---its running estimate of the long-run reward attainable from each state---based on observed rewards. Unlike a memoryless searcher, the agent carries forward what it has learned, so resetting does not merely shorten individual trajectories---it shapes how experience accumulates across training. Resetting thus acts through two channels. First, by truncating long trajectories, it can raise the rate at which the agent encounters rewards. Second, it can reshape how reward information propagates across the state space. Which channel dominates, and under what conditions, is the central question. 

Here, we show that stochastic resetting---a tunable, randomized counterpart to the episode-length truncation already common in reinforcement learning~\cite{pardo18}---accelerates reinforcement learning beyond its effect on first-passage time. In tabular Q-learning~\cite{watkins89,watkins92} on an open GridWorld, resetting speeds learning by more than it speeds the random search of an otherwise identical diffusive agent. Sharp (deterministic) resetting reaches a higher peak learning speedup than stochastic resetting, but over a narrower band of reset rates. On a stochastic cliff task (WindyCliff), resetting accelerates learning without reshaping the optimal policy, in contrast to the discount factor, which reshapes it directly. In a continuous-state benchmark (MountainCar~\cite{moore90}) solved with a deep Q-network (DQN)~\cite{mnih15}, resetting accelerates training when exploration is hard and rewards are sparse.

Together, these results identify stochastic resetting as a tunable mechanism for accelerating reinforcement learning, extending statistical-physics resetting theory from non-adaptive processes to systems that learn from experience.
\section*{Results}

\begin{figure}
    \centering
    \includegraphics[width=\columnwidth]{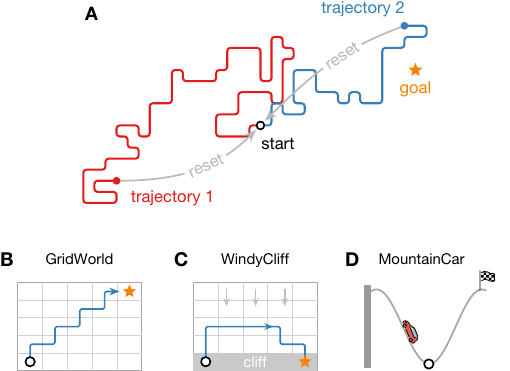}
    \caption{%
    \textbf{Competing effects of resetting across reinforcement-learning environments.}
    \textbf{(A)}~Two trajectories begin at the same start state (open circle). Resetting to the start (gray) shortens the remaining search for a trajectory that has wandered away from the goal, but lengthens it for one that has already moved closer; whether resetting accelerates search on average depends on the balance between these cases, set by the reset rate, the environment and the underlying process.
    \textbf{(B)}~GridWorld: a tabular agent learns a shortest path on an open square lattice.
    \textbf{(C)}~WindyCliff: stochastic wind and cliff penalties create a policy tradeoff from an externally imposed reset.
    \textbf{(D)}~MountainCar: an underpowered car must build momentum to reach the goal.
    In all environments, resetting returns the agent to the start during training without producing a value-function update or erasing learned values.%
    }
    \label{fig:schematic}
\end{figure}

We study resetting across three reinforcement-learning environments of increasing complexity: an open GridWorld and a stochastic cliff task (WindyCliff), both with tabular Q-learning, and a continuous-state MountainCar benchmark solved with a DQN (Fig~\ref{fig:schematic}B--D). The reset is an external intervention applied before action selection. Under the stochastic protocol, the agent returns to the start with a fixed reset probability at each training step; under the sharp protocol, it returns at fixed intervals matched to give the same mean reset rate. A reset shortens the remaining path to the goal when the agent has moved away from it and lengthens the path when the agent has moved closer (Fig~\ref{fig:schematic}A). Reset transitions produce no value update, and the Q-table or network weights persist; resetting modifies only the distribution of visited states. We evaluate the learned policy periodically in a greedy test episode (i.e., the agent always takes its currently best-valued action with exploration and resetting switched off) and record the \emph{test steps} needed to reach the goal. We track progress against \emph{cumulative training steps}, the running total of environment interactions, which dominate the cost of reinforcement learning. See Materials and Methods for algorithmic and hyperparameter details.

\begin{figure}
    \centering
    \includegraphics[width=\columnwidth]{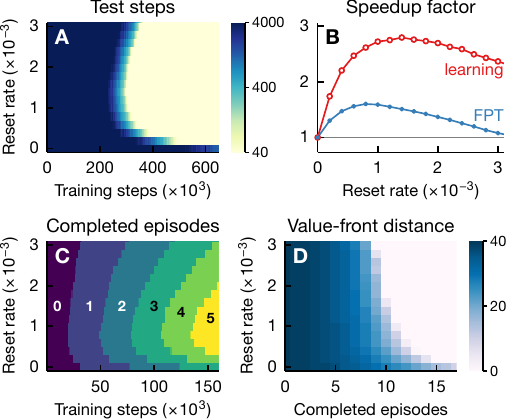}
    \caption{%
    \textbf{Stochastic resetting accelerates Q-learning on an open GridWorld, and the learning speedup exceeds the diffusive first-passage-time speedup.}
    \textbf{(A)}~Median test steps of the greedy policy as a function of training step and stochastic reset rate; color encodes test steps on a logarithmic scale, and pale cells mark convergence toward the 40-step optimum. Moderate reset rates reach the optimum earlier than the no-reset baseline.
    \textbf{(B)}~Median speedup relative to the no-reset baseline. The learning speedup---baseline convergence step divided by the convergence step at rate $r$---exceeds the diffusive first-passage-time (FPT) speedup, the analogous median ratio for an undirected random walker, across the useful range of reset rates.
    \textbf{(C)}~Median number of completed episodes during the first 160{,}000 training steps. Moderate reset rates discover the goal fastest, whereas high rates suppress discovery by truncating trajectories before arrival.
    \textbf{(D)}~Median value-front distance against completed episodes. Plotting against completed episodes isolates per-episode value propagation from the search clock; the residual separation across reset rates shows that resetting speeds value propagation independently of episode-completion rate. Unlike the episode-completion rate in (C), this effect strengthens monotonically with reset rate. 
    GridWorld of size $N\!=\!120$ with $\gamma\!=\!0.99$, $\alpha\!=\!0.0001$ and exploration rate $\varepsilon\!=\!1.0$; learning statistics use $2048$ matched seeds and the diffusive baseline $20{,}000$.%
    }
    \label{fig:gridworld_learning_dynamics}
\end{figure}

\subsection*{Resetting accelerates reinforcement learning beyond its effect on random search}
\label{sec:gridworld}
\begin{figure}[hbt!]
    \centering
    \includegraphics[width=\columnwidth]{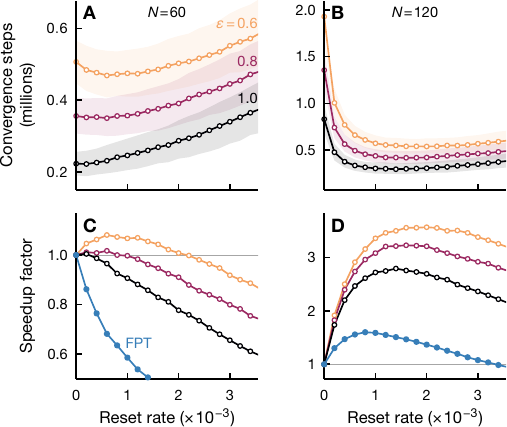}
    \caption{%
    \textbf{Resetting can speed learning beyond its effect on random search, and the separation is clearest in the larger GridWorld.}
    \textbf{(A)}~Median convergence step against stochastic reset rate for $N\!=\!60$ at exploration rates $\varepsilon\!=\!0.6$, $0.8$ and $1.0$.
    \textbf{(B)}~The same exploration sweep for $N\!=\!120$.
    \textbf{(C)}~Speedup factors for $N\!=\!60$; the learning speedup is compared with the diffusive FPT speedup.
    \textbf{(D)}~Speedup factors for $N\!=\!120$, where the learning speedups exceed the FPT speedups by a wide margin at intermediate reset rates.
    Hyperparameters as in Fig~\ref{fig:gridworld_learning_dynamics}, with the diffusive baseline over $20{,}000$ replicates.%
    }
    \label{fig:gridworld_beyond_search}
\end{figure}

Stochastic resetting accelerates Q-learning on the open GridWorld, and faster random search alone does not account for the gain. Figure~\ref{fig:gridworld_learning_dynamics}A shows a heatmap of median test steps over training step and reset rate. We see that moderate reset rates reach the optimum (40 steps) earlier than the no-reset baseline. We quantify this effect by the speedup---the training a no-reset agent needs to achieve a near-optimal policy divided by the training the resetting agent needs.
Fig~\ref{fig:gridworld_learning_dynamics}B compares the median speedups for learning and for the FPT of the equivalent diffusive search. The learning speedup is greater than the FPT speedup at every nonzero reset rate, and is maximized at a higher reset rate. We use the fastest-converging exploration rate, $\varepsilon\!=\!1$ (SI Fig~\ref{fig:gridworld_beyond_search}). Importantly, this choice holds the exploration policy fixed throughout training so that any speedup beyond the first-passage speedup reflects faster value propagation rather than improved exploration.

Two contributions are at work. Fig~\ref{fig:gridworld_learning_dynamics}C depicts the median number of completed episodes in the early phase of learning (up to the first 160{,}000 training steps). Moderate reset rates produce the highest episode-completion rate, whereas high rates too often truncate trajectories before they arrive. A faster completion rate alone, however, would reproduce only the FPT speedup. The gap in Fig~\ref{fig:gridworld_learning_dynamics}B shows that each completed episode is also more informative. Value propagates backward from the goal through successive Bellman updates~\cite{watkins92,sutton98}, and resetting prunes longer, indirect trajectories---the surviving shorter paths carry the value front closer to the start per episode. To isolate this from the episode-completion rate, Fig~\ref{fig:gridworld_learning_dynamics}D depicts the value-front distance, defined as the shortest distance from the start to any state with a nonzero value, against completed episodes rather than training steps. Resetting drives the value front toward the start substantially faster on this comparison, and the effect strengthens monotonically with the reset rate, unlike the episode-completion rate. Resetting accelerates learning via two channels, raising the episode-completion rate and making each completed episode more informative.

On a smaller grid (SI Fig~\ref{fig:gridworld_beyond_search}), resetting hurts both diffusive search and learning, but learning suffers less; the gap between learning and random-search speedups persists across grid sizes. Where resetting accelerates learning, learning speed depends nonmonotonically on the reset rate, with the maximum speedup at an intermediate value.

\begin{figure}
    \centering
    \includegraphics[width=\columnwidth]{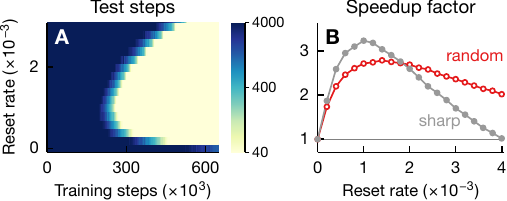}
    \caption{%
    \textbf{Sharp resetting yields a larger but narrower learning-speed optimum than stochastic resetting.}
    \textbf{(A)}~Median test steps of the greedy policy under sharp (deterministic) resetting, as a function of training step and equivalent reset rate $r\!=\!1/T$, on the same logarithmic scale as Fig~\ref{fig:gridworld_learning_dynamics}A.
    \textbf{(B)}~Median learning speedup relative to no reset for stochastic and sharp resetting. Sharp resetting reaches a larger peak speedup but is less forgiving, falling off more steeply at high reset rates.
    Same GridWorld and hyperparameters as Fig~\ref{fig:gridworld_learning_dynamics} ($N\!=\!120$, $\gamma\!=\!0.99$, $\alpha\!=\!0.0001$, $\varepsilon\!=\!1.0$; $2048$ seeds); sharp resetting returns the agent to the start every $T\!=\!\mathrm{round}(1/r)$ steps.%
    }
    \label{fig:gridworld_reset_modes}
\end{figure}

\subsection*{Sharp resetting yields a higher but narrower learning-speed optimum}
\label{sec:sharp-vs-stochastic}
The temporal structure of resets also shapes learning. In statistical physics, sharp resetting---in which the searcher returns to the start at fixed intervals---can achieve a lower minimum first-passage time than stochastic resetting~\cite{pal17,bhat16}. Sharp resetting is the direct analog of episode-length truncation, a heuristic widely used in reinforcement learning~\cite{pardo18}. Our results show that sharp resetting yields a higher peak learning speedup than stochastic resetting, but over a narrower range of reset rates (Fig~\ref{fig:gridworld_reset_modes}).

Fig~\ref{fig:gridworld_reset_modes} shows that both protocols speed up learning, with optima at slightly different reset rates. Above the optimum, the sharp-resetting speedup curve falls off steeply, whereas the stochastic curve degrades gradually. The tradeoff reflects how the two protocols affect long trajectories. Terminating every trajectory at a fixed interval deterministically removes the right tail of the trajectory-length distribution. Near the optimum, the removed trajectories are predominantly unproductive exploration, and eliminating them completely accelerates learning more than the partial removal under stochastic resetting. At higher reset rates, the same truncation discards trajectories that would have reached the goal, suppressing the episode-completion rate. Since the value front migrates toward the start during training, the best reset interval drifts. A fixed sharp reset interval is optimal for one distance, whereas stochastic resetting spans a distribution of intervals and is thus more forgiving.

\begin{figure}[hbt!]
    \centering
    \includegraphics[width=\columnwidth]{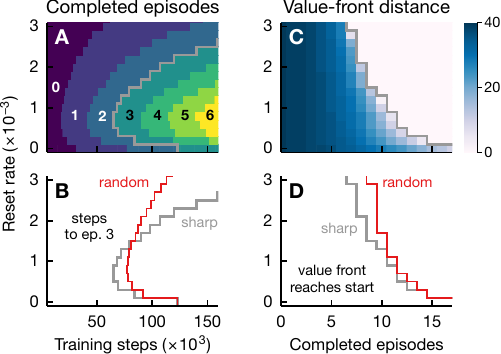}
    \caption{%
    \textbf{Sharp resetting changes the episode clock and the per-episode value-propagation clock in different ways.}
    \textbf{(A)}~Median completed episodes under sharp resetting as a function of training step and equivalent reset rate; the stepped overlay marks the training step at which the third episode is completed.
    \textbf{(B)}~Training steps needed to reach the third completed episode, for stochastic and sharp resetting. Sharp resetting reaches the third episode sooner at low and intermediate rates, but later at high rates, where fixed-interval truncation more often interrupts a trajectory before it arrives.
    \textbf{(C)}~Median value-front distance under sharp resetting as a function of completed episodes and reset rate; the stepped overlay marks the episode at which the value front reaches the start.
    \textbf{(D)}~Completed episodes needed for the value front to reach the start, for stochastic and sharp resetting; sharp resetting propagates value to the start in fewer episodes at every rate.
    Same GridWorld and hyperparameters as Fig~\ref{fig:gridworld_learning_dynamics}.%
    }
    \label{fig:gridworld_reset_modes_mechanism}
\end{figure}

SI Fig~\ref{fig:gridworld_reset_modes_mechanism} illustrates this mechanism by separating per-episode value propagation from the episode-completion rate. Across all reset rates, sharp resetting propagates the value front to the start in fewer completed episodes than stochastic resetting (SI Fig~\ref{fig:gridworld_reset_modes_mechanism}D; cf.\ Fig~\ref{fig:gridworld_learning_dynamics}D). The training-step picture is mixed. SI Fig~\ref{fig:gridworld_reset_modes_mechanism}B shows that sharp resetting reaches a given episode count in fewer training steps at low and intermediate reset rates, but in more training steps at high rates (cf.\ Fig~\ref{fig:gridworld_learning_dynamics}C). The high-rate collapse of the sharp-resetting speedup results from this reversal---at high reset rates, the loss of goal-reaching episodes outweighs the more efficient per-episode propagation.

\begin{figure}
    \centering
    \includegraphics[width=\columnwidth]{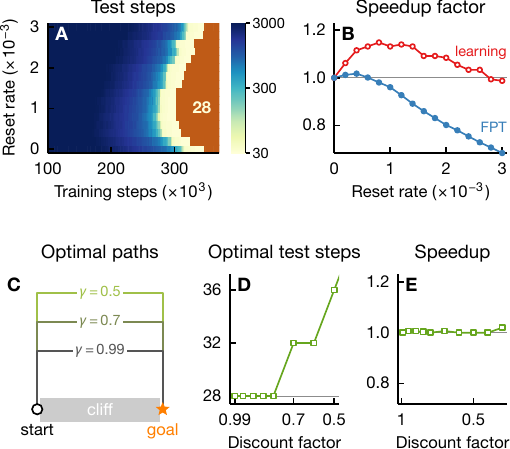}
    \caption{%
    \textbf{Resetting accelerates WindyCliff learning without changing the converged policy, whereas the discount factor reshapes the optimal policy.}
    \textbf{(A)}~Median test steps of the greedy policy at $\gamma\!=\!0.99$ as a function of training step and stochastic reset rate; the orange region marks convergence toward the $28$-step learned path.
    \textbf{(B)}~Reset learning speedup and diffusive FPT speedup relative to the no-reset baseline. The learning speedup exceeds the FPT speedup at every reset rate, and stays above one even where resetting slows the diffusive searcher.
    \textbf{(C)}~Dynamic-programming (DP) optimal paths for $\gamma\!=\!0.5$, $0.7$ and $0.99$; a lower discount selects a longer, safer route away from the cliff.
    \textbf{(D)}~DP-optimal no-wind path length as a function of discount, shortening as $\gamma$ increases.
    \textbf{(E)}~Discount-only learning speedup, which stays essentially flat---tuning $\gamma$ does not accelerate learning.
    WindyCliff of width $300$ and height $150$ with wind probability $p_w\!=\!0.005$ and strength $s_w\!=\!3$; reset panels use $\gamma\!=\!0.99$, $\alpha\!=\!0.5$, $\varepsilon\!=\!0.5$ and $600{,}000$ training steps over $2048$ seeds, with $20{,}000$ diffusive replicates. Panels D and E use a no-reset sweep over $\gamma$.%
    }
    \label{fig:windycliff_learning_dynamics}
\end{figure}

\subsection*{Resetting accelerates learning without altering the optimal policy}
\label{sec:windycliff}
Beyond the protocol comparison, a natural question is whether resetting acts on learning differently from the controls already standard in reinforcement learning. The most direct comparison is the discount factor~$\gamma$, which controls how strongly the agent weighs future rewards against immediate ones. We turn to WindyCliff, a stochastic cliff task on a $300\!\times\!150$ grid in which the start and goal lie on the bottom row, separated by a stretch of cliff cells. Wind pushes the agent downward with small probability after each action, so paths skirting the cliff risk falling, which incurs a large negative reward and teleports the agent back to the start. The optimal policy is sensitive to~$\gamma$. Smaller~$\gamma$ discounts the distant goal relative to the immediate cliff penalty and favors longer, cliff-avoidant routes, while larger~$\gamma$ favors shorter, riskier ones.

Resetting accelerates WindyCliff learning. As in GridWorld, at $\gamma\!=\!0.99$ the median test steps reach their floor earliest under intermediate reset rates (Fig~\ref{fig:windycliff_learning_dynamics}A), and the learning speedup exceeds the diffusive FPT speedup at every reset rate we study (Fig~\ref{fig:windycliff_learning_dynamics}B; cf.\ Fig~\ref{fig:gridworld_learning_dynamics}B). At high reset rates, resetting becomes detrimental to diffusive search in this geometry. The FPT speedup falls below one, while the learning speedup remains above one, indicating that faster random search does not account for the gain.

The discount factor, by contrast, modifies the optimum itself. Dynamic-programming (DP) solutions of WindyCliff show that smaller~$\gamma$ selects longer routes well away from the cliff, while larger~$\gamma$ selects shorter, riskier routes (Fig~\ref{fig:windycliff_learning_dynamics}C). The DP-optimal path length shortens monotonically with~$\gamma$ (Fig~\ref{fig:windycliff_learning_dynamics}D).

Tuning~$\gamma$ does not accelerate learning. The learning speedup across $\gamma$ values is essentially flat, in contrast to the strongly nonmonotonic dependence on reset rate (Fig~\ref{fig:windycliff_learning_dynamics}E; cf.\ Fig~\ref{fig:windycliff_learning_dynamics}B). Our results show that the discount factor reshapes the DP-optimal policy, while the reset rate accelerates learning without changing it.

\begin{figure}
    \centering
    \includegraphics[width=\columnwidth]{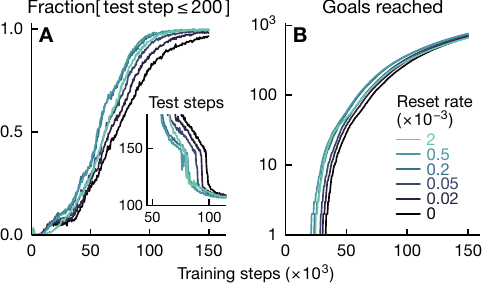}
    \caption{%
    \textbf{Stochastic resetting accelerates deep reinforcement learning in the extended MountainCar.}
    \textbf{(A)}~Fraction of replicates whose greedy-evaluation performance is at most $200$ steps, as a function of cumulative training steps for each reset rate (legend). Intermediate reset rates rise earlier and stay higher than the no-reset baseline. Inset: median greedy-evaluation steps.
    \textbf{(B)}~Median cumulative number of goals reached during training; intermediate reset rates raise the goal-encounter rate, while the highest rates interrupt the momentum-building trajectories needed to reach the goal.
    Extended MountainCar (left boundary $x_{\min}\!=\!-1.7$) under the sparse-positive reward ($+1$ at the goal, $0$ otherwise), solved with a deep Q-network at $\gamma\!=\!0.98$, learning rate $0.0001$ and final exploration $\varepsilon\!=\!0.1$; $150{,}000$ training steps and $2048$ replicates per reset rate.%
    }
    \label{fig:mountaincar}
\end{figure}

\subsection*{Resetting accelerates deep reinforcement learning}
\label{sec:mountaincar}
\begin{figure}[hbt!]
    \centering
    \includegraphics[width=\columnwidth]{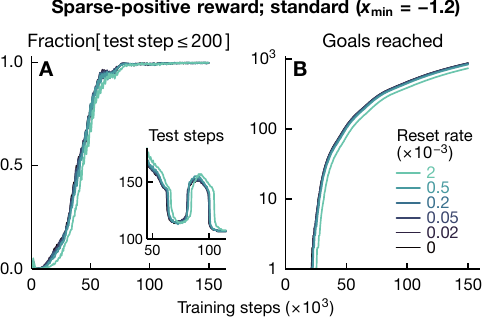}
    \caption{%
    \textbf{Standard MountainCar with a sparse-positive reward.}
    As Fig~\ref{fig:mountaincar}, but with the standard left boundary ($x_{\min}\!=\!-1.2$). The valley is shallow enough that the agent reliably discovers the goal unaided, so resetting offers no benefit.
    \textbf{(A)}~Fraction of replicates with greedy-evaluation performance at most $200$ steps; inset, median greedy-evaluation steps.
    \textbf{(B)}~Median cumulative goals reached.
    Hyperparameters as in Fig~\ref{fig:mountaincar}.%
    }
    \label{fig:mc-si-standard-sparse}
\end{figure}

To investigate stochastic resetting beyond tabular settings, we turn to the DQN algorithm applied to a modified MountainCar environment (Fig~\ref{fig:schematic}D), a continuous-state benchmark in which an underpowered car must escape a valley by building momentum. The left boundary is extended to $-1.7$, introducing a deep trap that makes unassisted goal discovery substantially harder, and the reward is sparse positive ($+1$ at the goal, $0$ elsewhere). We compare with a step-penalty reward and with the standard left boundary ($-1.2$) as controls (SI Figs~\ref{fig:mc-si-standard-sparse} and~\ref{fig:mc-si-standard-penalty}). For each reward scheme, we tune the learning rate, exploration schedule and target-update period to the best no-reset performance so the gains from resetting are improvements over an already-optimized baseline rather than artifacts of a weak one. Full details are provided in Materials and Methods.

In the extended environment under the sparse reward scheme, intermediate reset rates accelerate learning. The fraction of replicates achieving evaluation performance of at most 200 steps rises earlier and stays higher compared to the no-reset baseline (Fig~\ref{fig:mountaincar}A), and median evaluation steps drop faster (Fig~\ref{fig:mountaincar}A, inset). The mechanism is visible in the training dynamics. Resetting increases the rate at which the agent encounters the goal (Fig~\ref{fig:mountaincar}B), truncating unproductive excursions deep into the extended valley and returning the agent to the informative region near the goal. Too-high reset rates, however, interrupt trajectories so frequently that the agent cannot build the momentum needed to reach the goal, impairing both the encounter rate and the learned policy.

In this setting, the visible mechanism is the goal-encounter rate (Fig~\ref{fig:mountaincar}B) rather than the per-episode value propagation observed in GridWorld (Fig~\ref{fig:gridworld_learning_dynamics}D). In both cases, resetting shortens uninformative trajectories, but the bottleneck it alleviates differs. Here, discovering the goal is the limiting factor. The MountainCar result therefore probes the scope of the reset intervention under function approximation rather than the value-propagation channel isolated in the tabular settings.

\begin{figure}[hbt!]
    \centering
    \includegraphics[width=\columnwidth]{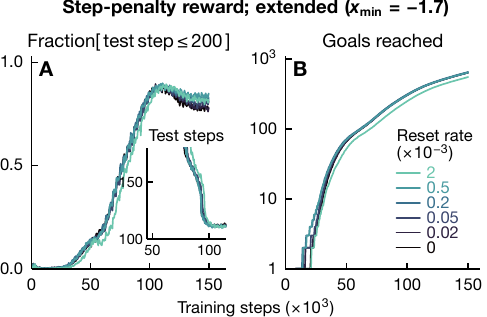}
    \caption{%
    \textbf{Extended MountainCar with a step-penalty reward.}
    As Fig~\ref{fig:mountaincar}, but with a step-penalty reward ($-1$ per step, $0$ at the goal) on the extended environment ($x_{\min}\!=\!-1.7$). The penalty supplies a gradient signal on every transition, so exploration is no longer the bottleneck: learning curves for different reset rates largely overlap and the highest rates modestly impair performance.
    \textbf{(A)}~Fraction of replicates with greedy-evaluation performance at most $200$ steps; inset, median greedy-evaluation steps.
    \textbf{(B)}~Median cumulative goals reached.
    Step-penalty hyperparameters: learning rate $0.004$, final exploration $\varepsilon\!=\!0.07$; $150{,}000$ training steps and $2048$ replicates per reset rate.%
    }
    \label{fig:mc-si-extended-penalty}
\end{figure}

\begin{figure}[h!]
    \centering
    \includegraphics[width=\columnwidth]{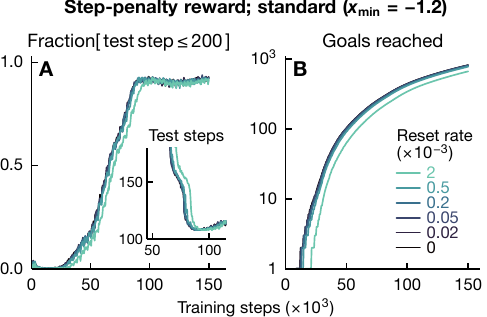}
    \caption{%
    \textbf{Standard MountainCar with a step-penalty reward.}
    As Fig~\ref{fig:mc-si-extended-penalty}, but with the standard left boundary ($x_{\min}\!=\!-1.2$). With both a dense penalty signal and easy goal discovery, resetting again offers no benefit.
    \textbf{(A)}~Fraction of replicates with greedy-evaluation performance at most $200$ steps; inset, median greedy-evaluation steps.
    \textbf{(B)}~Median cumulative goals reached.
    Step-penalty hyperparameters as in Fig~\ref{fig:mc-si-extended-penalty}.%
    }
    \label{fig:mc-si-standard-penalty}
\end{figure}

Under the step-penalty reward ($-1$ per step, $0$ at the goal), the penalty provides a gradient signal on every transition regardless of whether the agent reaches the goal; exploration is no longer the primary bottleneck. Learning curves for different reset rates largely overlap, and the largest rates modestly impair performance (SI Fig~\ref{fig:mc-si-extended-penalty}). When the left boundary is at its standard value ($-1.2$), the valley is shallow enough that the agent reliably discovers the goal without assistance; resetting offers no benefit under either reward scheme (SI Figs~\ref{fig:mc-si-standard-sparse} and~\ref{fig:mc-si-standard-penalty}). Our results demonstrate that stochastic resetting can accelerate deep RL when the environment poses a hard exploration problem and the reward is sparse.
\section*{Discussion}
We show that stochastic resetting accelerates reinforcement learning in all three settings considered. In the tabular Q-learning experiments, the learning speedup exceeds the FPT speedup of a diffusive searcher under the same protocol; in WindyCliff the gap persists even at reset rates that slow the diffusive searcher itself. We find that sharp resetting reaches a higher peak learning speedup than stochastic resetting, but over a narrower range of reset rates. In WindyCliff the discount factor reshapes the DP-optimal policy while resetting leaves it unchanged. In MountainCar with a deep Q-network, resetting accelerates training when exploration is hard and rewards are sparse. Our results identify resetting as an intervention on the distribution of trajectories, distinct from changes to the reward signal or the update rule.

Existing stochastic resetting theory addresses static processes, whose dynamics do not change between resets~\cite{evans11,evans20,pal17}. A learning agent, by contrast, updates its policy with experience, making the dynamics that resetting interrupts fundamentally nonstationary. This opens two channels through which resetting can accelerate learning: raising the rate at which the agent encounters rewards by truncating long trajectories, and reshaping how reward information propagates through the state space. Our results reveal that both channels are at play, with their relative weight depending on specific details of the environment. In GridWorld, the value front advances toward the start substantially faster per completed episode under resetting (Fig~\ref{fig:gridworld_learning_dynamics}D); this channel acts independently of the episode-completion rate. WindyCliff corroborates this picture. At high reset rates, resetting slows the diffusive searcher while still accelerating learning (Fig~\ref{fig:windycliff_learning_dynamics}B); faster random search cannot account for the gain. In MountainCar, by contrast, the visible mechanism is the goal-encounter rate (Fig~\ref{fig:mountaincar}B), because finding the goal at all is the limiting factor.

These results position stochastic resetting as a distinct intervention among approaches to accelerating reinforcement learning. Unlike intrinsic reward bonuses, uncertainty-driven exploration, temporally persistent actions and mechanisms that return agents to previously visited states before expanding outward~\cite{burda18,ecoffet21,eysenbach18,osband16,pathak19,mouret15}, resetting modifies only the temporal structure of experience, through a single control parameter. This simplicity also comes with tradeoffs, namely that the scope of resetting is limited: in environments with deceptive rewards or partial observability, where progress relies on identifying informative behaviors rather than abandoning unproductive trajectories, richer exploration methods remain necessary.

Resetting has begun to enter machine learning along several lines: a first-passage framework for choosing perturbations that speed up neural-network training, stochastic resets that mitigate noisy-label overfitting and reinforcement-learning agents that learn when to reset during target search~\cite{meir25,bae25,munoz25}. The fixed, external protocol we study is complementary, isolating the effect of resetting on the temporal structure of experience.

Several extensions of the protocol invite further study. 
First, a state-dependent reset rate would target truncation to states where unproductive trajectories are most likely, exploiting the value-front mechanism. 
Second, restart distributions could replace the fixed home with states of high learned value, coupling resetting to the geometry of the value landscape. Adapting the start-state distribution is itself an established technique. Reverse-curriculum methods grow the set of start states outward from the goal, exploiting the same backward value propagation through a learned, nonstationary distribution rather than a fixed external rate~\cite{florensa17}. A related direction has recently emerged in language-model post-training, where resetting to a self-localized reasoning error and resampling continuations sharpens credit assignment~\cite{samanta26}.
Third, the sharp-resetting limit invites a closer look at episode-length truncation as deployed in practice~\cite{pardo18}; the narrower optimum of sharp resetting (Fig~\ref{fig:gridworld_reset_modes}) suggests stochastic resetting may be more robust when the optimal interval is unknown a priori. Optimizing the temporal structure of resetting---for instance, a reset schedule that adapts as the value front advances---could improve on either fixed protocol.

More generally, this work points toward future directions connecting stochastic resetting theory and reinforcement learning to natural instances of restart-like behavior, such as central-place foraging~\cite{orions79} and kinetic proofreading~\cite{hopfield74,ninio75,ravasio26,rotbart15}. Of particular interest for learning are novelty-induced resets in neural systems during exploration~\cite{park21}. Whether the idealized resetting process studied here captures the phenomena observed in these systems remains an open question, but the connections we identify may offer a theoretical foothold.

\section*{Materials and Methods}
\subsection*{Tabular grid environments}

GridWorld is an open $N \times N$ lattice with four cardinal-direction actions clipped at the boundaries. The start is at $(N/2-10,\, N/2-10)$, the goal at $(N/2+10,\, N/2+10)$, separated by Manhattan distance $40$, the optimal path length. The reward is $+1$ at the goal and zero elsewhere. The main results use $N = 120$ (Fig~\ref{fig:schematic}B); SI Figs~\ref{fig:gridworld_robustness} and \ref{fig:gridworld_beyond_search} also include $N = 60$.

WindyCliff is a $300 \times 150$ grid adapted from the Gymnasium CliffWalking task~\cite{towers25} (Fig~\ref{fig:schematic}C). The start and goal lie on the bottom row at columns $140$ and $160$; the remaining bottom-row cells form a cliff. Reaching the goal yields $+10$ and terminates the episode. Stepping onto a cliff cell yields $-100$ and immediately teleports the agent to the start, with the resulting transition included in the Q-update; this teleportation is non-terminal and distinct from stochastic resetting. All other transitions yield zero reward. Stochastic wind pushes the agent downward by $s_w = 3$ cells with probability $p_w = 0.005$ after each action.

\subsection*{Q-learning update}

Both grid environments use tabular Q-learning with an $\varepsilon$-greedy policy~\cite{watkins89,watkins92,sutton98}. With the action-value function $Q(s,a)$ initialized to zero, each transition $(s,\, a,\, r_t,\, s')$ produces the Bellman update
\begin{equation}
    Q(s,a) \leftarrow Q(s,a) + \alpha \left[r_t + \gamma \max_{a'} Q(s',a') - Q(s,a)\right],
\end{equation}
with learning rate $\alpha$ and discount factor $\gamma$. The agent acts uniformly at random with probability $\varepsilon$ and greedily otherwise. Both $\alpha$ and $\varepsilon$ are held fixed throughout training. Episodes begin at the start state and terminate when the agent reaches the goal.

For GridWorld, we use $\gamma = 0.99$, $\alpha = 10^{-4}$ and $\varepsilon = 1.0$ in the main figures (Figs~\ref{fig:gridworld_learning_dynamics} and~\ref{fig:gridworld_reset_modes}) and in SI Fig~\ref{fig:gridworld_reset_modes_mechanism}; SI Fig~\ref{fig:gridworld_beyond_search} sweeps $\varepsilon \in \{0.6,\, 0.8,\, 1.0\}$. Training proceeds for $2 \times 10^{6}$ steps at $N = 120$ and $10^{6}$ steps at $N = 60$, with the greedy policy evaluated every $1{,}000$ steps via a single $5{,}000$-step rollout. For WindyCliff, we use $\gamma = 0.99$, $\alpha = 0.5$ and $\varepsilon = 0.5$, training for $6 \times 10^{5}$ steps with greedy evaluation every $2{,}000$ steps; each checkpoint averages $16$ rollouts of horizon $5{,}000$ to absorb wind variance. The greedy policy is evaluated with exploration and resetting disabled, recording steps to the goal. We report medians across $2{,}048$ matched-seed replicates.

\subsection*{MountainCar environment}

The MountainCar dynamics follow the standard Gymnasium specification~\cite{towers25}, with two modifications. The initial state is fixed at $(\text{position},\, \text{velocity}) = (-0.5,\, 0)$ rather than sampled uniformly in $[-0.6,\, -0.4]$, and the left boundary is set to $-1.7$ (``extended'') or $-1.2$ (``standard''); the extended valley is a deep trap that makes unassisted goal discovery substantially harder. Two reward schemes are studied: sparse positive ($+1$ at the goal, zero otherwise) and step penalty ($-1$ per step, zero at the goal). The main result (Fig~\ref{fig:mountaincar}) uses the sparse-positive extended condition.

We approximate the action-value function with a deep Q-network~\cite{mnih15}: a two-hidden-layer perceptron of $256$ ReLU units per layer mapping the two-dimensional observation to Q-values for three discrete actions. Training uses experience replay (buffer $10{,}000$; batch size $128$; learning begins after $10^{3}$ steps, then $8$ gradient steps every $16$ environment steps), a target network with periodic hard updates, smooth $L_1$ loss and the Adam optimizer~\cite{kingma15}. The exploration rate $\varepsilon$ is linearly annealed from $1.0$ to $\varepsilon_\mathrm{end}$ over the first $20\%$ of training. The discount factor is $\gamma = 0.98$ for both reward schemes. The sparse-positive scheme uses learning rate $10^{-4}$, $\varepsilon_\mathrm{end} = 0.10$ and a target-network update every $400$ steps; the step-penalty scheme uses learning rate $4 \times 10^{-3}$, $\varepsilon_\mathrm{end} = 0.07$ and a target-network update every $600$ steps. We train for $150{,}000$ steps, evaluating the greedy policy every $500$ steps with a $10{,}000$-step horizon, and report statistics across $2{,}048$ independent replicates.

\subsection*{Stochastic and sharp resetting protocols}

Stochastic resetting is implemented identically across all three environments. Before action selection at each training step, the agent is returned to the designated start state with probability $r$ fixed throughout training, independent of state and action. The reset transition produces no value-function update; the Q-table or network weights persist, so resetting modifies only the distribution of training trajectories. In MountainCar, replay-buffer value targets are computed from the underlying (non-reset) transition, avoiding spurious bootstraps across resets. Since resetting alters neither the reward structure nor the value-function definition, it leaves the tabular optimal policy unchanged.

The deterministic, or sharp, limit of this protocol returns the agent to the start every $T$ steps, with $T = \mathrm{round}(1/r)$ matching the mean reset rate of the stochastic protocol at probability $r$. Sharp resetting is closely related to episode-length truncation, a common practical heuristic in reinforcement learning~\cite{pardo18}. The two protocols are compared in Fig~\ref{fig:gridworld_reset_modes} and SI Fig~\ref{fig:gridworld_reset_modes_mechanism}; all other figures use the stochastic protocol.

\subsection*{Convergence criterion and learning speedup}

\begin{figure}[hbt!]
    \centering
    \includegraphics[width=\columnwidth]{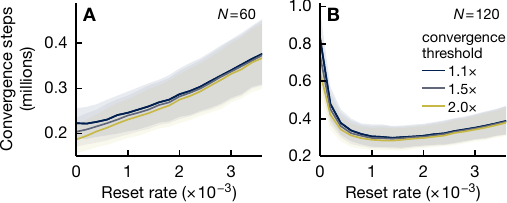}
    \caption{%
    \textbf{The GridWorld learning-speed conclusion is robust to the near-optimality threshold used to define convergence.}
    \textbf{(A)}~Median convergence step against stochastic reset rate for $N\!=\!60$, using convergence thresholds of $1.1$, $1.5$ and $2.0$ times the $40$-step optimal path.
    \textbf{(B)}~The same threshold comparison for $N\!=\!120$; the intermediate-reset optimum persists across thresholds.
    Hyperparameters as in Fig~\ref{fig:gridworld_learning_dynamics} ($\gamma\!=\!0.99$, $\alpha\!=\!0.0001$, $\varepsilon\!=\!1.0$).%
    }
    \label{fig:gridworld_robustness}
\end{figure}

We measure learning progress by the first greedy-evaluation checkpoint at which the median episode length falls within a near-optimal envelope of the dynamic-programming-optimal path length. For GridWorld, the envelope is $1.1\times$ the optimal length of $40$ steps (a $10\%$ margin); SI Fig~\ref{fig:gridworld_robustness} shows the conclusions are unchanged when the threshold is widened to $1.5\times$ and $2\times$. For WindyCliff, stochastic wind enlarges the variance of the evaluation curve, and we instead use a $1.5\times$ envelope of the DP-optimal path length at each $\gamma$. For MountainCar, neither criterion is well suited to the continuous-state setting; the main metric (Fig~\ref{fig:mountaincar}A) is the fraction of replicates whose greedy episode length falls below $200$ steps as a function of training step.

The learning speedup at reset rate $r$ is the ratio of the no-reset median convergence step to the median convergence step at rate $r$, taken over $2{,}048$ matched-seed replicates. For the WindyCliff discount-factor sweep (Fig~\ref{fig:windycliff_learning_dynamics}E), the same definition is applied to no-reset ($r = 0$) runs at $\gamma \in \{0.3,\, 0.4,\, 0.5,\, 0.6,\, 0.7,\, 0.8,\, 0.85,\, 0.9,\, 0.95,\, 0.99\}$, with the $1.5\times$ threshold recomputed at each $\gamma$ from the corresponding DP-optimal path length.

\subsection*{Value-front distance}

To track how reward information propagates across the state space, we define the value-front distance as the smallest Manhattan distance from the start to any state whose greedy value differs from its initial value of zero. At training step $t$, with start state $s_0$,
\begin{equation}
    d_\mathrm{vf}(t) = \min\left\{\, \lVert s - s_0 \rVert_1 \;:\; \max_a Q_t(s,a) \neq 0 \,\right\}.
\end{equation}
Before any state has nonzero value, we set $d_\mathrm{vf}$ to the start-goal Manhattan distance, $40$, equivalent to initializing the front at the goal.
With the sparse-positive reward and $Q \equiv 0$ initialization, $Q(s,a)$ first becomes nonzero on the state-action pair one step before the goal, and the resulting ``value front'' propagates back toward the start through successive Bellman updates. The quantity $d_\mathrm{vf}(t)$ thus tracks the spatial reach of learned values.

Fig~\ref{fig:gridworld_learning_dynamics}D and SI Fig~\ref{fig:gridworld_reset_modes_mechanism} plot $d_\mathrm{vf}$ against completed episodes rather than training step, isolating per-episode value propagation from the episode-completion rate. The alignment is constructed per replicate: for each integer episode count $k$, we record $d_\mathrm{vf}$ at the first checkpoint that registers $k$ completed episodes, then take the median across the $2{,}048$ replicates.

\subsection*{Diffusive first-passage baselines}

To separate learning gains from generic reductions in random-search time, we compare each tabular result against a memoryless agent on the same environment. The diffusive baseline takes uniformly random actions at every step and is subject to the same stochastic resetting protocol and reset-rate grid as the learning agent. For WindyCliff, the diffusive walker is exposed to identical wind and cliff teleportation but has no value function to update. For each $(N,\, r)$ we record the first-passage time (FPT) to the goal across $20{,}000$ independent trials and report the median; the FPT speedup is the ratio of the no-reset median FPT to the median FPT at rate $r$. This baseline is used in Fig~\ref{fig:gridworld_learning_dynamics}B (GridWorld, $N = 120$) and Fig~\ref{fig:windycliff_learning_dynamics}B (WindyCliff); the $N = 60$ counterpart appears in SI Fig~\ref{fig:gridworld_beyond_search}.

\subsection*{Dynamic-programming baselines}

For WindyCliff we compute the optimal action-value function $Q^*(s,a)$ and greedy policy $\pi^*(s) = \arg\max_a Q^*(s,a)$ by iterating the Bellman expectation
\begin{equation}
    Q_{k+1}(s,a) = \mathbb{E}\left[r(s,a,s') + \gamma \max_{a'} Q_k(s',a')\right]
\end{equation}
over stochastic wind outcomes, until the maximum change in the value function falls below $10^{-12}$. The DP-optimal path length at each $\gamma$ is the deterministic rollout length of $\pi^*$ from the start state, used both as the WindyCliff convergence threshold and as the speedup reference  (the DP-optimal path length in Fig~\ref{fig:windycliff_learning_dynamics}D). The three optimal paths in Fig~\ref{fig:windycliff_learning_dynamics}C correspond to $\gamma = 0.5$, $0.7$ and $0.99$.

\begin{acknowledgments}
JZ acknowledges funding from Stanford University through the Graduate Program in Biophysics. VN acknowledges research funds from the University of Sydney. The research of VN was supported in part by grants from the NSF (DMS-2235451) and Simons Foundation (MPS-NITMB-00005320) to the NSF-Simons National Institute for Theory and Mathematics in Biology (NITMB). DJS was partially supported by a Simons Fellowship in the MMLS, a Sloan Fellowship and the National Science Foundation, through the Center for the Physics of Biological Function (PHY-1734030). VN and DJS also acknowledge funding from Coefficient Giving.
\end{acknowledgments}

\onecolumngrid
\newpage
\section*{Supplementary Figures}
\twocolumngrid

\newpage

\newpage

\eject

\end{document}